# Enhance the Image: Super Resolution using Artificial Intelligence in MRI

A book chapter in *Machine Learning in MRI: From methods to clinical translation*

Ziyu Li[1,2], Zihan Li[1], Haoxiang Li[1], Qiuyun Fan[3], Karla L. Miller[2], Wenchuan Wu[2], Akshay S. Chaudhari[4,5], Qiyuan Tian[1]

[1]School of Biomedical Engineering, Tsinghua University, Beijing, China;
[2]Wellcome Centre for Integrative Neuroimaging, FMRIB, Nuffield Department of Clinical Neurosciences, University of Oxford, Oxford, UK;
[3]Academy of Medical Engineering and Translational Medicine, Medical College, Tianjin University, Tianjin, China;
[4]Department of Radiology, Stanford University, Stanford, California, USA;
[5]Department of Biomedical Data Science, Stanford University, Stanford, California, USA.

Abstract:
This chapter provides an overview of deep learning techniques for improving the spatial resolution of MRI, ranging from convolutional neural networks, generative adversarial networks, to more advanced models including transformers, diffusion models, and implicit neural representations. Our exploration extends beyond the methodologies to scrutinize the impact of super-resolved images on clinical and neuroscientific assessments. We also cover various practical topics such as network architectures, image evaluation metrics, network loss functions, and training data specifics—including downsampling methods for simulating low-resolution images and dataset selection. Finally, we discuss existing challenges and potential future directions regarding the feasibility and reliability of deep learning-based MRI super-resolution, with the aim to facilitate its wider adoption to benefit various clinical and neuroscientific applications.

Keywords:
Single-image super-resolution, deep learning, convolutional neural network, generative adversarial network, transformer, diffusion model, implicit neural representation, loss function, transfer learning, uncertainty estimation.

1. Introduction

MRI with higher spatial resolution provides more detailed insights into the structure and function of living human bodies non-invasively, which is highly desirable for accurate clinical diagnosis and image analysis. The spatial resolution of MRI is characterized by in-plane and through-plane resolutions (Fig. 1). The in-plane spatial resolution along a dimension ($\Delta x$) depends on the maximal extent of the k-space ($k_x^{\max}$):

$$\Delta x = \frac{1}{2k_x^{\max}}, \qquad (1)$$

where a broader coverage of k-space $k_x^{\max}$ corresponds to a higher $\frac{1}{\Delta x}$ and therefore a higher in-plane resolution (Fig. 1a). On the other hand, the through-plane resolution, also referred to as slice thickness, is determined differently for 2D and 3D imaging. In 2D imaging, the slice thickness is defined by the full width at half maximum (FWHM) of the slice-selection radiofrequency (RF) pulse profile. In 3D imaging, the slice-selection direction is encoded by another phase encoding gradient. Consequently, the through-plane resolution is determined similarly to the in-plane resolution by the maximal extent of the k-space along slice-selection direction as in Eq. 1.

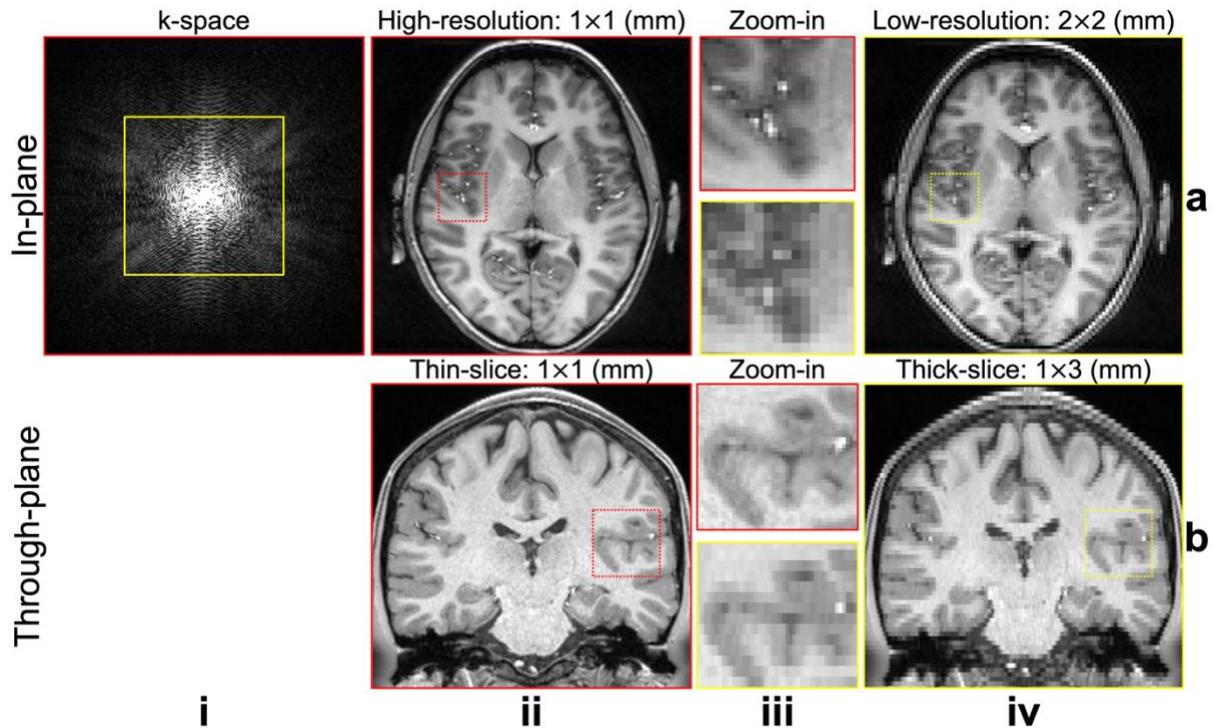

**Figure 1. Spatial resolution of MRI.** The in-plane resolution is dictated by the k-space coverage, and a larger k-space coverage brings higher spatial resolution (a). The slice thickness is determined by the slice-selective RF pulse for 2D imaging, and by k-space extent along slice-selection direction for 3D imaging (b).

However, higher spatial resolution in MRI is achieved at the cost of prolonged scan times. Depending on the scheme used for image formation, the pursuit of higher in-plane resolution through a broader k-space coverage generally necessitates an extended readout time, more phase-encoding steps, and/or more excitations. Regarding the through-plane resolution, for 2D imaging, a reduction in slice thickness escalates the number of slices to acquire, leading to increased scan time. More importantly, the ability of 2D imaging to resolve very thin slice (e.g., 1 mm or thinner) is intrinsically limited by the sharpness of RF profiles. Consequently, the demand for achieving slender slice thickness and superior, isotropic resolution often mandates the adoption of 3D imaging that is often more time-consuming due to an additional phase-encoding dimension and substantially increased phase-encoding steps. Moreover, to compensate the decreased SNR for high-resolution scans with smaller voxel size, averaging multiple repetitions of data is often required. The prolonged acquisition imposes challenges such as increased scan costs, subject discomfort, susceptibility to motion artifacts, and impracticality for subjects intolerant to extended scans, such as certain populations of patients and children.

Image super-resolution technology offers a promising venue to harness the advantages of higher resolution for a short scan. It is a well-established problem which aims to restore a high-resolution image from a given low-resolution input [1], often referred to as single-image super-resolution and formulated as an ill-posed inverse problem that incorporates appropriate priors. Early efforts in computer vision indicated that priors derived from high-resolution example patches effectively enhance low-resolution patches sharing similar textures [2, 3]. In the context of MRI, various priors, including high-resolution example-based priors [4], non-local patches [5], self-similarity [5, 6], sparse encoding [7], total variation and low-rankness [8] have been proposed. However, these methods only demonstrated moderate resolution improvement due to the intrinsically limited high-frequency information in peripheral k-space and the difficulty in solving the nonlinear, ill-posed problem. An alternative category of MRI super-resolution methods uses multiple acquired images with different slice shifts [9], FOV orientations [10, 11], or RF encodings [12-14]. Van Reeth et al. comprehensively reviewed this type of approaches [15]. Despite their appealing performance especially for improving the through-plane resolutions, they necessitate extended scan times to acquire multiple images. The primary focus of this chapter is on single-image super-resolution techniques that do not necessitate additional scan time.

Recent strides in machine learning and deep learning have exhibited significant potential for achieving substantial resolution improvement from a single input image. It has been demonstrated that even a shallow 3-layer convolutional neural network (CNN) can outperform state-of-the-art traditional methods, such as sparse encoding, for super-resolving 2D natural images [16]. Increasing the depth of CNNs [17] and employing advanced training strategies, such as adversarial training [18], can further enhance their performance. Impressively, these deep learning methods are also faster and easier to deploy once the networks are trained. Their success in natural image super-resolution, coupled with these advantages, positions them as a promising method for achieving high-fidelity MRI super-resolution without the need of additional scan time.

In this chapter, we offer a comprehensive overview of current deep learning-based MRI super-resolution techniques. Our discussion encompasses not only the methodologies but also the impact of super-resolved images on clinical and neuroscientific assessments. Delving into practical considerations, we cover various aspects such as network architectures, image evaluation metrics, network loss functions, and pertinent training considerations—ranging from methods for simulating low-resolution images to dataset selection. Finally, we evaluate existing challenges and propose potential future directions for deep learning-based MRI super-resolution, consolidating insights for researchers and practitioners in this evolving field.

## 2. Deep learning-based MRI super-resolution

### 2.1. Standard CNN

After achieving notable success in natural image super-resolution within the field of computer vision, CNNs have found widespread applications in MRI super-resolution. The fundamental framework of CNN-based MRI super-resolution is depicted in Fig. 2.

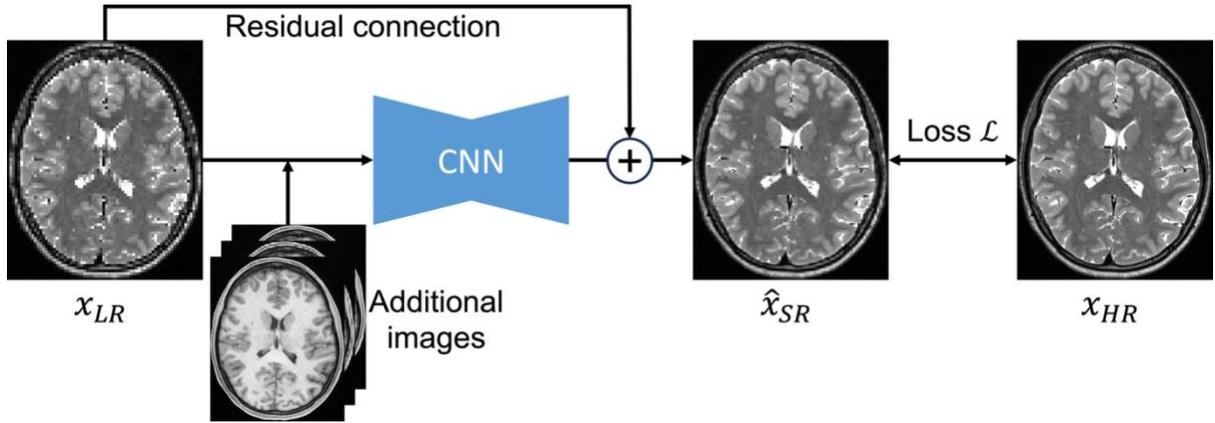

**Figure 2. CNN-based MRI super-resolution.** A basic framework for super-resolution MRI using CNN is demonstrated. The CNN is optimized by minimizing the loss between the high-resolution target $x_{HR}$ and the super-resolved image $\hat{x}_{SR}$ generated from the low-resolution input $x_{LR}$. Residual learning strategy can be adopted for faster convergence and boosted performance. Additional images that are available can be also included to leverage redundant information.

Similar to its application in natural image super-resolution, CNN in MRI is employed to transform low-resolution input with thick slice thickness and/or low in-plane resolution into super-resolved images. Low-resolution inputs are often interpolated to match the dimensions of the high-resolution target using methods like trilinear or cubic spline interpolation. The optimization process involves minimizing the discrepancy, quantified through metrics such as mean squared error (MSE, L2-loss) or mean absolute error (MAE, L1-loss), between the network's output and the high-resolution training target. Taking MSE as an example loss function, the training objective is formulated as:

$$\min_{\theta} \mathcal{L}_{L2}(\theta) = \left|\left|\hat{x}_{SR} - x_{HR}\right|\right|_2^2 = \left|\left|f_{\theta}(x_{LR}) - x_{HR}\right|\right|_2^2, \quad (2)$$

where $x_{LR}$, $\hat{x}_{SR}$, and $x_{HR}$ are the interpolated low-resolution input, the CNN-synthesized super-resolved image, and target high-resolution reference, respectively. $f_{\theta}$ is the CNN parametrized by $\theta$.

Given the 3D nature of MRI data, CNNs with 3D convolutional kernels are commonly used to exploit redundancy along all three dimensions. Notably, a shallow 3-layer 3D CNN [19] has demonstrated enhanced performance for brain MRI super-resolution when compared to traditional methods such as non-local upsampling, total variation and low-rank-based super-resolution, and their 2D CNN counterparts.

The efficacy of deeper networks has been established, with the addition of more layers further enhancing performance [20]. Drawing inspiration from ResNet [21] that introduced residual connections to facilitate the training of deep CNNs and unleash their full potential for nonlinear mapping in natural image classification, studies have confirmed the beneficial impact of residual connections on both natural image [17] and MRI super-resolution [22-24]. Intuitively, CNNs with residual connections can focus on learning to estimate high-frequency components, requiring less information and leading to enhanced performance, particularly considering the number of CNN parameters and available training resources. Specifically, instead of directly generating $x_{HR}$, residual learning estimates the difference between $x_{LR}$ and $x_{HR}$:

$$\min_\theta \left\| f_\theta^R(x_{LR}) - (x_{HR} - x_{LR}) \right\|_2^2, \tag{3}$$

where $f_\theta^R$ is the CNN parametrized by $\theta$ optimized using residual learning strategy. The final super-resolved output $\hat{x}_{SR}$ can be obtained by adding the interpolated low-resolution input to the network output:

$$\hat{x}_{SR} = x_{LR} + f_\theta^R(x_{LR}). \tag{4}$$

Moreover, the inclusion of complementary images has been identified as a means to potentially enhance performance. For brain MRI, Zeng et al. showed that including high-resolution T1-weighted images contributes to the better performance of T2-weighted images super-resolution [25]. Lyu et al. proposed a progressive network that also incorporates high-resolution images of proton density and T1-weighted contrast for T2-weighted image super-resolution [26]. In cardiac imaging, the incorporation of multiple time frames into the input has been shown to marginally improve the accuracy of super-resolution [27]. Since these complementary images are routinely acquired in practice, their integration does not increase the scan time. This approach of leveraging additional images to enhance super-resolution performance has also been explored in other medical imaging modalities. For example, the inclusion of high-resolution MRI has been found beneficial for deep learning-based positron emission tomography (PET) super-resolution [28].

2.2. Generative adversarial network (GAN)

While standard CNNs trained using metrics like MSE or MAE have demonstrated promising results in MRI super-resolution, with high peak signal-to-noise ratios (PSNR) compared to native high-resolution images, these super-resolved images do not necessarily exhibit satisfying visual quality. The limitation lies in the use of pixel-wise loss functions such as MSE, which face challenges in effectively addressing the inherent uncertainty associated with recovering lost high-frequency texture details. The minimization of MSE tends to favor pixel-wise averages of plausible solutions, leading to outcomes that are often oversmoothed and, consequently, exhibit limited perceptual quality [18].

GAN has proven to be an effectively solution to this problem. GAN consists of a generator and a discriminator [29] (Fig. 3). For MRI super-resolution, the generator super-resolves the input low-resolution images. CNNs discussed in Section 2.1 can be considered as a generator. Meanwhile, the discriminator's role is to classify an image as either real or generated. The generator and the discriminator are trained alternatively to compete against each other. As training progresses, the discriminator becomes highly proficient in distinguishing between real and generated images. Simultaneously, the generator refines its ability to produce realistic-looking images based on the feedback from the discriminator. Eventually, the discriminator becomes an adept classifier, while the generator excels at generating images that closely resemble real ones, as guided by the learned criteria of the discriminator.

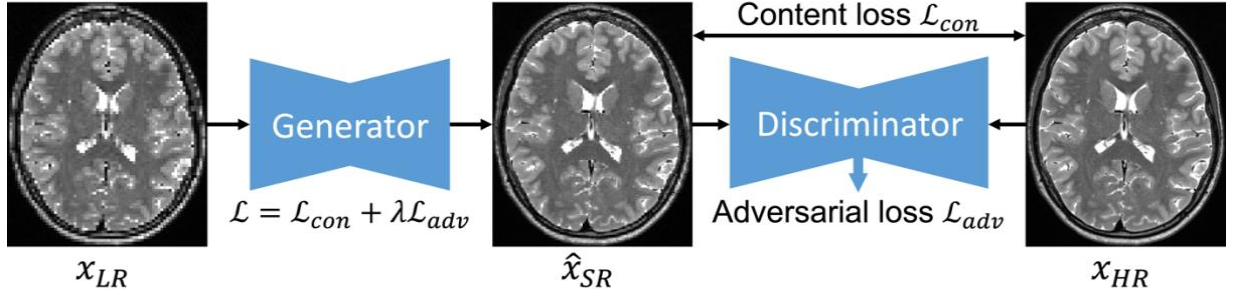

**Figure 3. GAN for MRI super-resolution.** The GAN comprises a generator which super-resolves the input low-resolution image $x_{LR}$, and a discriminator which discerns between the generated image $\hat{x}_{SR}$ and the original high-resolution image $x_{HR}$. The generator and discriminator are trained to compete against each other. This adversarial training dynamic aims to converge towards the generation of realistic high-resolution images indistinguishable from the native high-resolution counterparts.

Mathematically, when the generator is trained, the discriminator is fixed, and the optimizer tries to solve:

$$\min_{\theta_G} \mathcal{L}^G_{con} + \lambda \mathcal{L}^G_{adv}, \tag{5}$$

where $\theta_G$ is the generator network parameters, $\mathcal{L}^G_{con}$ is the content loss between the generated super-resolved image and the high-resolution target (e.g., MSE as in Eqs. 1 and 2), and $\mathcal{L}^G_{adv}$ is the adversarial loss derived from the discriminator. This loss reflects the probability, as determined by the discriminator, that a super-resolved image could be classified as an actual high-resolution image. $\lambda$ is the hyperparameter which determines the contribution of the adversarial loss.

The adversarial loss can be defined as:

$$\mathcal{L}^G_{adv} = -\log D_{\theta_D}(\hat{x}_{SR}) = -\log D_{\theta_D}\left(G_{\theta_G}(x_{LR})\right), \tag{6}$$

where $G_{\theta_G}$ and $D_{\theta_D}$ denote the generator and discriminator parametrized by $\theta_G$ and $\theta_D$ respectively. $\theta_D$ is fixed during the training of the generator. $D_{\theta_D}(\hat{x}_{SR})$ represents the probability assigned by the discriminator to the classification of a super-resolved image as a genuine high-resolution image.

During the training of the discriminator, $\theta_G$ is fixed. The training process tries to optimize the parameters of the discriminator ($\theta_D$) such that it could accurately discriminate the generated and actual images. Using binary cross-entropy loss which can classify generated and target images, as originally proposed for GAN[29], the training objective of the discriminator can be formulated as:

$$\min_{\theta_D} \mathcal{L}^D = -\left(\log D_{\theta_D}(x_{HR}) - \log D_{\theta_D}(\hat{x}_{SR})\right) = -\left(\log D_{\theta_D}(x_{HR}) - \log D_{\theta_D}\left(G_{\theta_G}(x_{LR})\right)\right). \quad (7)$$

In practice, GANs with binary cross-entropy discriminator loss are notoriously challenging to train, often suffering from problems such as gradient vanishing, model collapsing, non-convergence etc. Regularization techniques (e.g., spectral normalization [30]) as well as other variants of GANs with different loss function (e.g., least-square GAN [31], Wasserstein GAN [32, 33]) have been proposed to stabilize the training.

Studies leverage GAN to improve the visual quality of the super-resolved MRI. For example, in brain MRI, a 3D Wasserstein GAN has been proposed for T1-weighted structural image super-resolution, which produces sharper and more realistic textures compared to its counterpart trained without adversarial loss (i.e., $\lambda$=0 in Eq. 5) [34, 35]. Furthermore, GAN has proven beneficial for simultaneous super-resolution and geometric distortion correction in diffusion-weighted images [36]. In knee MRI, an ensembled GAN has been proposed to transform images obtained from multiple conventional super-resolution methods into a more accurate super-resolved image with improved image fidelity [37].

2.3. Transformer

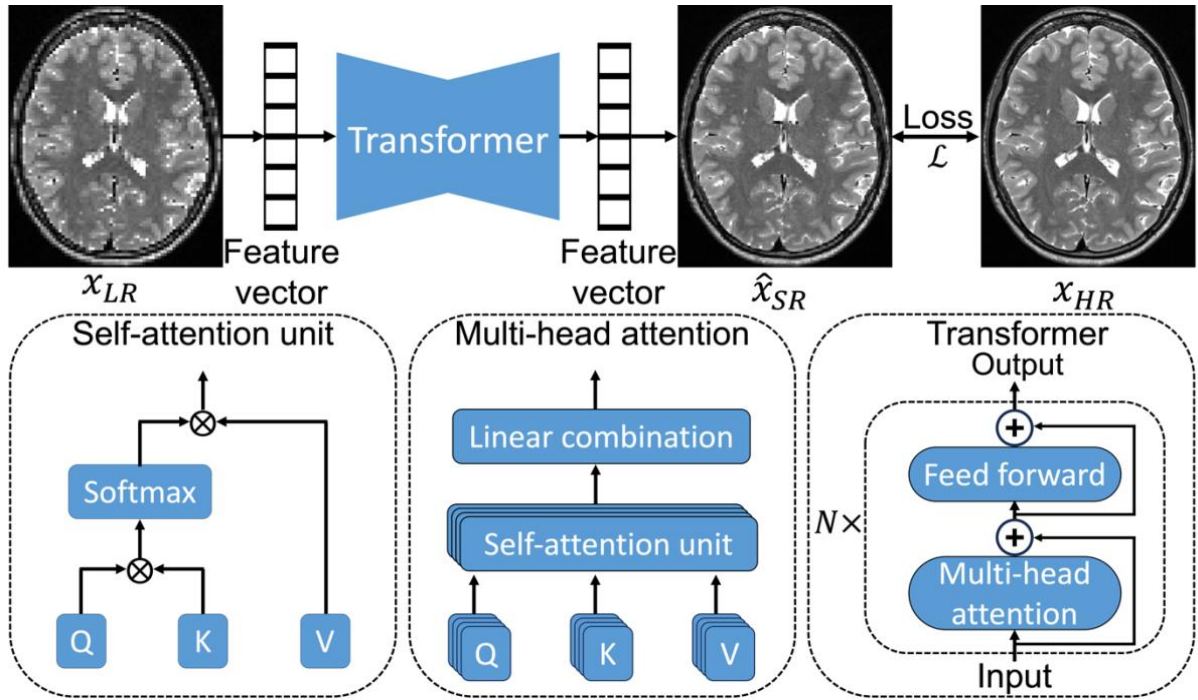

**Figure 4. Transformer framework for MRI super-resolution.** The transformer operates on the feature vectors extracted from the low-resolution input $x_{LR}$ and produces output feature vectors from which the super-resolved image $\hat{x}_{SR}$ can be generated. A self-attention unit is illustrated, where Q, K, V denote the query, key, and value matrices, respectively, and ⊗ represents the inner product operation. Multiple self-attention units can be stacked in parallel to form a multi-head attention unit. The transformer comprises N multi-head attention and feed forward blocks with residual connections.

An inherent limitation of CNNs is that their receptive field is constrained by kernel size and network architecture, which hinders their ability to capture long-range information for MRI super-resolution. Transformers have emerged as a promising architecture to address this challenge by modeling long-range dependencies within the image using the self-attention mechanism [38, 39]. The self-attention mechanism operates by transforming an input vector (e.g., a flattened patch of an image) into three vectors: the query vector $q$, the key vector $k$, and the value vector $v$. The vectors from all inputs are then stacked to form matrices $Q$, $K$, and $V$. This allows for the calculation of the attention function between all combinations of input vectors using an inner product and a Softmax function (Fig. 4) to indicate the importance of each input vector to each output. Essentially, the key $k$ identifies features, the query $q$ specifies the feature of interest, and the value $v$ is the response corresponding to a matching query-key pair.

Mathematically, given an input sequence of vectors $X$, the self-attention mechanism calculates the output feature vector $Y$ following:

$$Y = \text{softmax}\left(\frac{QK^T}{\sqrt{d_k}}\right)V = \text{softmax}\left(\frac{(W_Q X)(W_K X)^T}{\sqrt{d_k}}\right)(W_V X), \tag{8}$$

where $d_k$ is the dimension of the key vector $k$, and $W_Q$, $W_K$, and $W_V$ are learnable weight matrices for query, key, and value projections, respectively. Scaling the inner product of $QK^T$ with $1/\sqrt{d_k}$ is beneficial when the dimension of key vector is large [38]. Instead of performing a single attention function with multi-dimensional queries, keys, and values, it has been found beneficial to linearly project queries, keys, and values and parallelly perform self-attention function (i.e., having multiple $W_Q$, $W_K$, and $W_V$), known as multi-head attention (Fig. 4). The transformer comprises multiple multi-head attention and feed-forward blocks with residual connections to boost performance (Fig. 4). Additionally, the information about the position of each input/output element within the sequence can be leveraged using positional encoding, which is typically achieved by adding vectors to the sequence, where each vector encodes the relative position of the element within the sequence [38].

After initially being proposed for natural language processing tasks [38], transformers have found applications in vision tasks as well. For instance, the Vision Transformer (ViT) [40] flattens 2D patches to obtain vectors, which are then input to a transformer for image classification. However, pure transformers like ViT have not shown significant performance improvement compared to CNNs for vision tasks, possibly due to a limited ability to leverage local information [39]. For image super-resolution, the integration of convolutional layers with transformers emerges as a promising solution [41], where convolutional layers can be used to extract local features from input images as well as to reconstruct output images from features.

Transformers have demonstrated promise in MRI super-resolution, often incorporating auxiliary contrasts for enhanced performance. For example, Feng et al. employed a multi-modality transformer for super-resolution and reconstruction of brain and knee MRI, which introduces a cross-attention module to fully exploit information in each modality, outperforming CNN-based methods [42]. Fang et al. proposed a transformer for multi-modality MRI super-resolution, leveraging inter-modality correlation as well as high-frequency structure priors [43]. Li et al. introduced a transformer-based multi-scale network for multi-contrast MRI super-resolution [44]. Despite improved performance compared to CNNs, these studies also report increased model complexity and potential larger training data requirements for transformer-based models.

2.4. Diffusion model

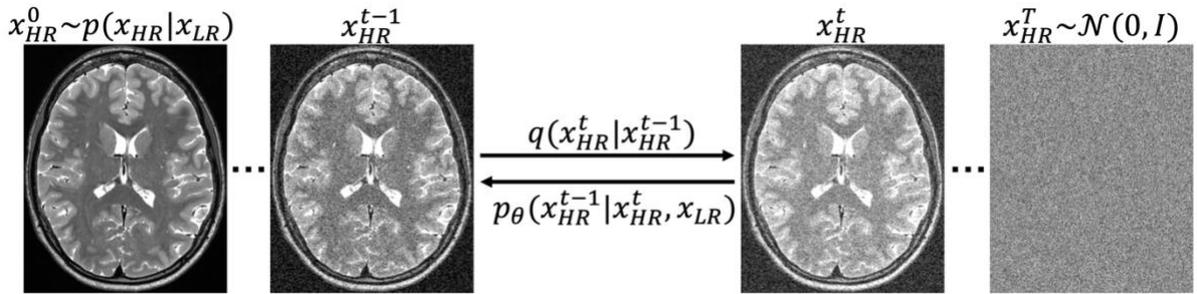

**Figure 5. Forward and reverse diffusion process.** The forward diffusion process $q$ (left to right) gradually adds Gaussian noise to the target high-resolution image $x_{HR}$. The reverse process $p$ (right to left) iteratively denoises $x_{HR}$, conditioned on a source low-resolution image $x_{LR}$, which is not shown.

Diffusion models, inspired by principles from non-equilibrium thermodynamics [45], establish a sequential diffusion process to gradually introduce random noise to the data. They then learn to reverse this process, ultimately reconstructing the original data from the noise [45, 46]. Diffusion models address the instability often associated with GANs in synthesizing high-fidelity images and have proven effective in generating realistic, high-quality natural images [46]. In the forward Markovian diffusion process $q$ gradually adds noise to the high-resolution image $x_{HR}^0$ over $T$ iterations (Fig. 5), ultimately generating a noise map which serves as a latent representation of the original data:

$$q(x_{HR}^{1:T}|x_{HR}^0) = \prod_{t=1}^{T} q(x_{HR}^t|x_{HR}^{t-1}), \qquad (9)$$

$$q(x_{HR}^t|x_{HR}^{t-1}) = \mathcal{N}(x_{HR}^t|\sqrt{\alpha_t}x_{HR}^{t-1}, (1-\alpha_t)I), \qquad (10)$$

where $\alpha_t \in (0,1)$ is the hyperparameter determining the variance of the added noise and $\mathcal{N}(0,I)$ is the normal distribution. $x_{HR}^{t-1}$ is attenuated by $\sqrt{\alpha_t}$ to ensure the random variable remains bounded when $t \to \infty$. To shorten the calculation time of $x_{HR}^t$, the reparameterization trick which sets $\bar{\alpha}_t := \prod_{s=1}^{t} \alpha_s$ will allow the sampling of $x_{HR}^t$ at any arbitrary time step from the following distribution [46]:

$$q(x_{HR}^t|x_{HR}^0) = \mathcal{N}(x_{HR}^t|\sqrt{\bar{\alpha}_t}x_{HR}^0, (1-\bar{\alpha}_t)I). \qquad (11)$$

The reverse process aims to recover the high-resolution image by gradually removing the noise ($p$ in Fig. 5). Without conditioning this will lead to the generation of random high-resolution images with similar distribution as the training data. For image super-resolution, we leverage the low-resolution image as conditioning [47, 48] to recover the super-resolved image:

$$p_\theta(x_{HR}^{0:T}|x_{LR}) = p(x_{HR}^T) \prod_{t=1}^{T} p_\theta(x_{HR}^{t-1}|x_{HR}^t, x_{LR}), \qquad (12)$$

$$p_\theta(x_{HR}^{t-1}|x_{HR}^t, x_{LR}) = \mathcal{N}\big(x_{HR}^{t-1}|\mu_\theta(x_{HR}^t, x_{LR}, t), \Sigma_\theta(x_{HR}^t, x_{LR}, t)\big), \qquad (13)$$

where $x_{HR}^T \sim \mathcal{N}(0, I)$. Essentially, the training of the diffusion model estimates the noise distribution $\mathcal{N}(\mu_\theta, \Sigma_\theta)$ for given timestep $t$, noisy image $x_{HR}^t$, and the low-resolution conditioning $x_{LR}$ using a neural network parametrized by $\theta$. Each training step takes a gradient descent step on:

$$\nabla_\theta \left\| \epsilon - \epsilon_\theta(x_{HR}^t, x_{LR}, t) \right\|_2^2 = \nabla_\theta \left\| \epsilon - \epsilon_\theta\big(\sqrt{\bar{\alpha}_t} x_{HR}^0 + (1 - \bar{\alpha}_t)\epsilon, x_{LR}, t\big) \right\|_2^2, \qquad (14)$$

where $\epsilon \sim \mathcal{N}(0, I)$. In practice, In practice, U-Net can be employed for learning the distribution, and the conditioning on $x_{LR}$ can be achieved by concatenating $x_{LR}$ to $x_{HR}^t$ along the channel dimension [47].

$T$ networks are trained to reverse $x_{HR}^T$ back to $x_{HR}^0$. During inference, $x_{HR}^{t-1}$ is calculated from:

$$x_{HR}^{t-1} = \frac{1}{\sqrt{\alpha_t}} \left( x_{HR}^t - \frac{1 - \alpha_t}{\sqrt{1 - \bar{\alpha}_t}} \epsilon_\theta(x_{HR}^t, x_{LR}, t) \right) + \sigma_t z, \qquad (15)$$

where $z \sim \mathcal{N}(0, I)$ when $t > 1$ (otherwise $z = 0$) and $\sigma_t^2 I = \Sigma_\theta(x_{HR}^t, x_{LR}, t)$. In practice $\sigma_t$ can be set to $\sqrt{1 - \alpha_t}$ [46, 47].

Diffusion models have proven effective in synthesizing high-fidelity, high-resolution MRI from low-resolution inputs. Wu et al. demonstrated the model's capability in synthesizing realistic-looking brain T1-weighted and diffusion-weighted MRI from 64× down-sampled low-resolution input [49]. Wang et al. [50] utilized a denoising diffusion implicit model [51] for optimal latent representations of low-resolution thick-slice brain MRI and employed a brain image synthesis diffusion model [52] pretrained on UK Biobank data to recover high-resolution brain images from the latent representations. Chung et al. [53] proposed a score-based diffusion model [54] which denoises knee and liver MRI and super-resolves the denoised images. The super-resolution is achieved using the same diffusion model by applying a data consistency term which takes blurring into account.

2.5. Implicit neural representation (INR)

As opposed to conventional pixel-based representation, INR aims to obtain a continuous representations of images, which theoretically allows the accurate generation of images in arbitrary resolutions by mapping any continuous image coordinates to the corresponding signal [55, 56].

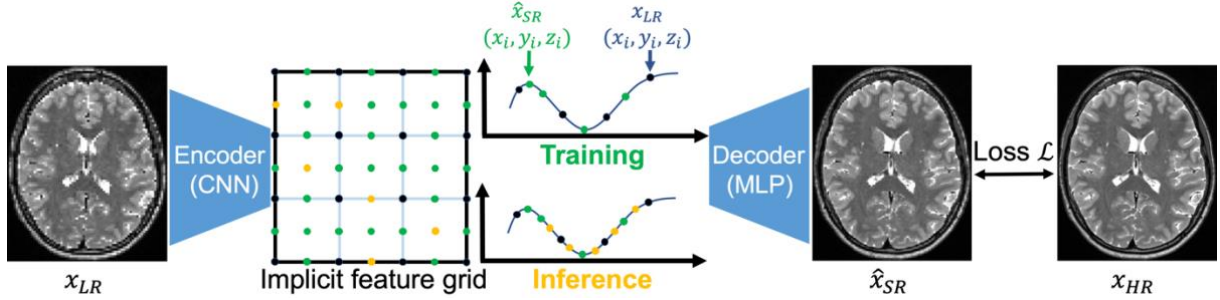

**Figure 6. Implicit neural representation for MRI super-resolution.** The input low-resolution image $x_{LR}$ is encoded as an implicit feature map using a convolutional neural network (CNN). The super-resolved image $\hat{x}_{SR}$ is reconstructed by finding the intensity of each voxel according to the coordinate and the implicit features using the decoder (often implemented as a multilayer perceptron (MLP)). The encoder and decoder are jointly optimized by minimizing the difference between $\hat{x}_{SR}$ and the target high-resolution image $x_{HR}$.

For image super-resolution with INR (Fig. 6), the representation can be learned using an encoder, typically implemented as a CNN to capture local information [56, 57]:

$$v = E_\varphi(x), \tag{16}$$

where $x$ is the input image, $v$ is the obtained feature map, and $E$ is the encoding network parametrized by $\varphi$.

Subsequently, high-resolution images can be reconstructed from the encoded implicit features, which spatially correspond to the object. The decoding step is often executed voxel by voxel through a multilayer perceptron (i.e., fully-connected neural network), taking the desired coordinate and corresponding feature as inputs:

$$\hat{x}_q = \mathcal{F}_\theta(v_q, q), \tag{17}$$

where $q$ is the image coordinate which can take arbitrary values and does not need to be integers, $\hat{x}_q$ and $v_q$ are the network estimated image intensity and feature at $q$, respectively, and $\mathcal{F}_\theta$ is the decoder network parametrized by $\theta$. For the query of $v_q$ with non-integer $q$, interpolation (e.g., nearest neighbor [56], trilinear [58]) can be applied to the feature map $v$. The spatial coordinate $q$ requires normalization for low- and high-resolution images.

The encoder and decoder can be optimized simultaneously by solving:

$$\min_{\varphi,\theta} \left\|\hat{x}_{SR,q} - x_{HR,q}\right\|_2^2 = \left\|\mathcal{F}_\theta\big(E_\varphi(x_{LR})_q, q\big) - x_{HR,q}\right\|_2^2. \tag{18}$$

The training process aims to fit a continuous implicit function that links spatial location to image intensity, enabling the acquisition of intensity at any desired location during inference (Fig. 6).

INR's superior performance for MRI super-resolution has been shown in various studies. Wu et al. leveraged INR for the super-resolution of thick-slice T1-weighted brain MRI, yielding high-fidelity, high-resolution brain images in both simulations and prospective in-vivo experiments [59]. They further extended their method to achieve robust arbitrary scale super-resolution, producing images with improved quality and more accurate brain region segmentations compared to CNN-based methods [58]. Li et al. showed that INR super-resolved brain quantitative susceptibility mapping (QSM) and T1-weighted images benefit the localization and delineation of human pedunculopontine nucleus [60]. Xu et al. proposed an INR-based framework which facilitates 3D volume reconstruction from thick-slice acquisition of fetal MRI, demonstrating improved image quality and shortened processing time compared to conventional methods [61].

3. Clinical and neuroscientific applications

These deep learning-based MRI super-resolution techniques hold significant promise in enhancing image quality while reducing scan time, making them valuable tools across various clinical and neuroscientific applications (Fig. 7).

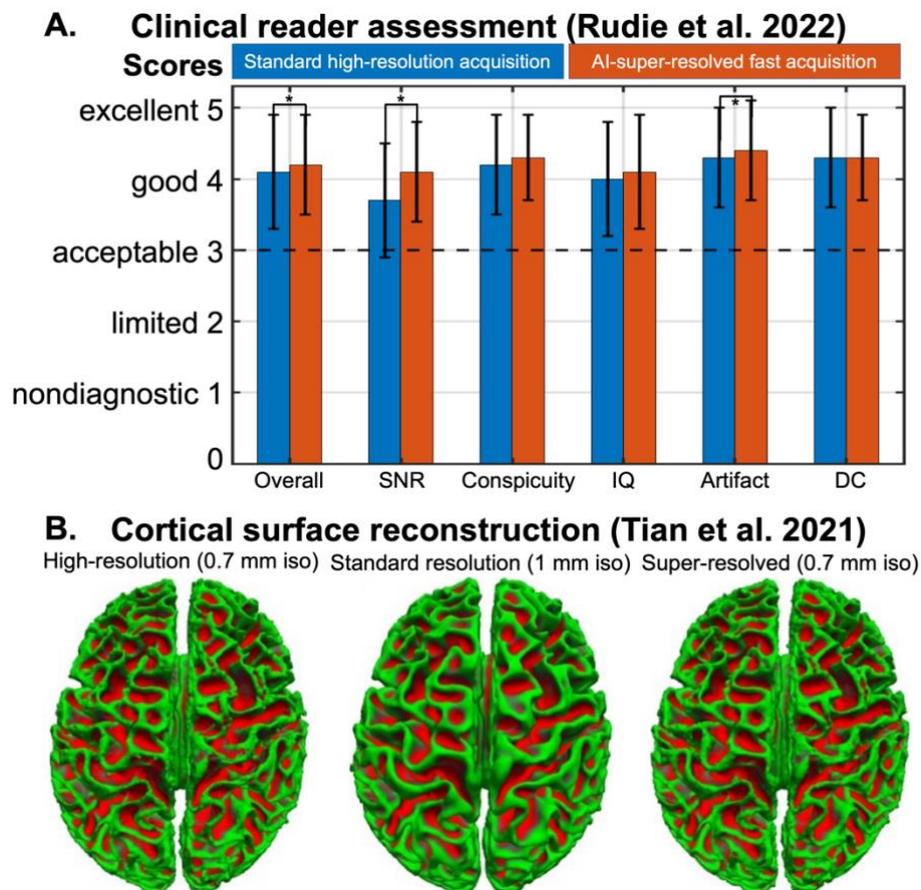

**Figure 7. Example clinical and neuroscientific applications of deep learning-based MRI super-resolution.** Rudie et al. conducted a comprehensive clinical assessment of brain MRI from fast acquisition with deep learning super-resolution and standard, longer high-resolution acquisition, with regard to overall evaluation, SNR, anatomical conspicuity, image quality (IQ), artifact [62]. The scores for T2-weighted FLAIR images are shown and comparisons with statistical significance are marked with asterisks (A). Tian et al. demonstrated deep learning-based super-resolution effectively improved the accuracy of cortical surface reconstruction and thickness estimation [24] (B).

A number of clinical assessments have underscored the clinical benefits of deep learning-based MRI super-resolution. For example, Rudie et al. [62] conducted a systematic clinical evaluation of deep learning super-resolved brain MRI. Four board-certified neuroradiologists participated in the study, assessing 3D T1-weighted pre-contrast, T1-weighted post-contrast, and T2-weighted FLAIR clinical brain MRI scans from 32 patients. The images were acquired through both standard high-resolution and a faster acquisition with lower resolution (~45% shorter scan time), which was subsequently super-resolved using a pre-trained U-Net. Evaluation criteria included SNR, anatomical conspicuity, image quality, artifact, and diagnostic confidence (Fig. 7A). Despite the ~45% reduction in scan time, the deep learning-enhanced images exhibited no significant compromise in image quality. On the contrary, they surpassed standard acquisition, particularly in terms of SNR and image artifact reduction. The authors attributed this improvement to the denoising effect of the network trained on large samples and the reduction in motion artifacts from the faster acquisition. Chaudhari et al. proposed and clinically evaluated "DeepResolve", a deep learning super-resolution method for knee MRI [22, 63]. This approach utilized a CNN with a residual learning strategy to map tricubic-interpolated thick-slice low-resolution images, obtained from 2D acquisition, to 3D high-resolution images acquired with a longer scan time. Initial assessments involved a reader study conducted by two radiologists who evaluated interpolated low-resolution, super-resolved, and native high-resolution images. The evaluation revealed a preference among radiographers for super-resolved images over interpolated low-resolution images in terms of image contrast, sharpness, SNR, artifacts, and overall quality [22]. They further extended their evaluation to qualitatively and quantitatively demonstrate that DeepResolve minimally biases cartilage and osteophyte biomarkers with superior image quality. Super-resolved images achieved reader scores, cartilage segmentation, and osteophyte detection comparable to native high-resolution images, substantially outperforming interpolated low-resolution images [63]. Masutani et al. demonstrated that for cardiac MRI, CNN super-resolved images exhibited similar ventricular volumetry and ejection fraction to native high-resolution images, with no statistical significance, supporting their use in clinical cardiac imaging [27].

Deep learning super-resolution techniques have also shown great potential for advancing neuroscientific applications. Along this line of research, Tian et al. utilized a CNN (i.e., VDSR [17]) to generate T1-weighted anatomical images at submillimeter isotropic resolution from standard T1-weighted images at 1 mm isotropic resolution. The resulting images exhibited superior gray-white contrast, improved cortical surface reconstruction, and more accurate cortical thickness estimation compared to interpolated standard 1 mm images (Fig. 7B). This development holds great promise in depicting neurological development, degeneration, and pathology, opening up new possibilities for cortical morphometry applications by significantly reducing scan times (from ~1 hour to less than 5 minutes) [24]. Iglesias et al. leveraged a 3D U-Net trained on synthetic data to achieve joint super-resolution and synthesis of 1 mm isotropic MP-RAGE volumes, allowing FreeSurfer-type [64-66] analyses such as cortical surface reconstruction and brain region segmentation. The method reliably estimated hippocampal volumes and right hemisphere thickness, maintained significance between Alzheimer's Disease and control groups, and proved valuable in registration for CNN-synthesized volumes super-resolved from thick-slice acquisition [67]. Li et al. employed a 3D U-Net to convert diffusion data to higher-resolution anatomical data, enabling various neuroscientific diffusion analyses, such as region-specific analysis, cortical surface-based analysis, and tractography [68]. Liu et al. also validated the reliability of CNN-super-resolved T1-weighted images in various neuroimaging tasks, including region-specific analysis, brain morphometry analysis, and structural covariance analysis [69]. It has also been demonstrated that CNN-based super-resolution for diffusion MRI reveals more detailed and accurate depiction of fiber distribution and orientation and enables more accurate tractography [36, 70-72].

The benefits of CNN-based MRI super-resolution for clinical and neuroscientific applications have been comprehensively validated. The super-resolved images are generally preferred by radiologists and can provide more accurate neuroscientific analyses compared to the low-resolution images. However, for more recent advances such as transformers, diffusion models, and INR-based methods, comprehensive evaluations and assessments are still limited, highlighting a potential direction for future studies.

## 4. Practical considerations

### 4.1. Network architecture

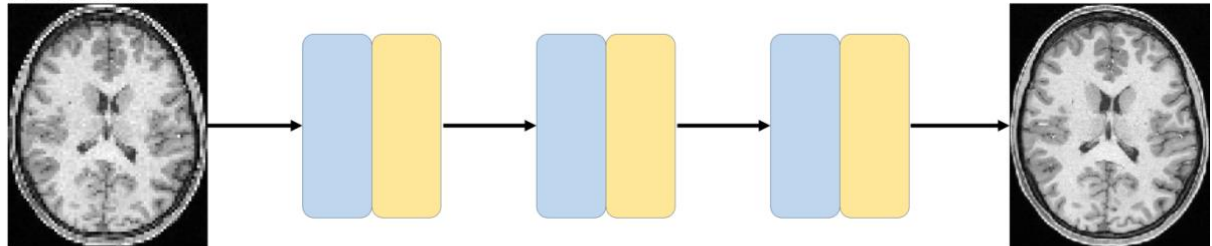
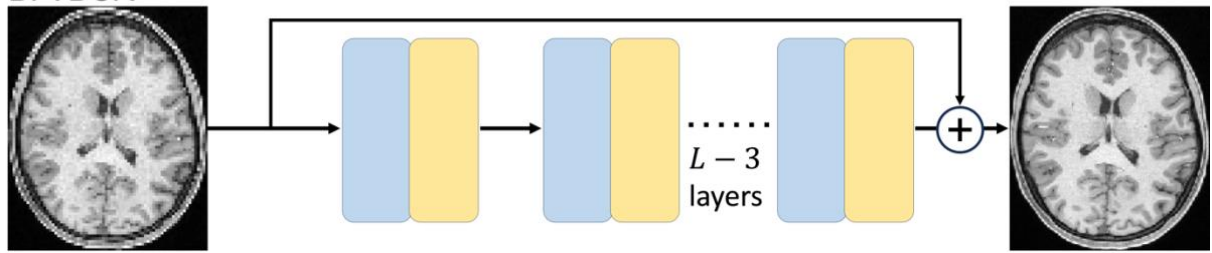
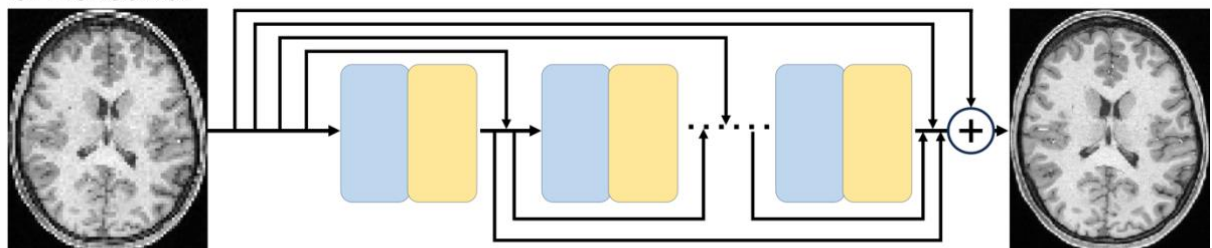
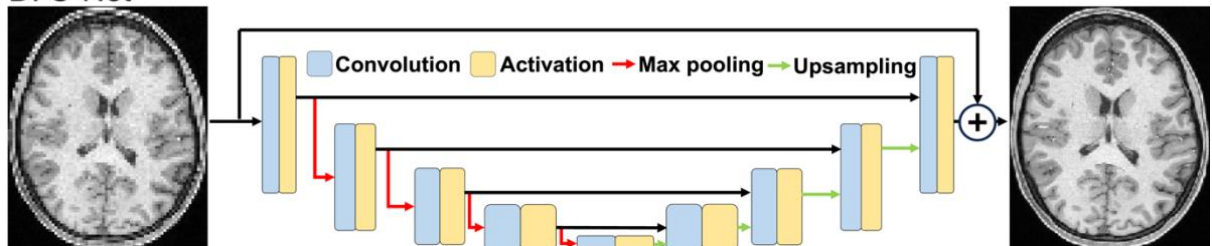

**Figure 8. Network architecture.** The architectures of several widely used networks for image super-resolution including SRCNN (A), VDSR (B), DenseNet (C), and U-Net (D) are illustrated. L denotes the total number of layers for VDSR (L=20 in [17]).

The application of CNN in image super-resolution started with a simple 3-layer architecture (i.e., SRCNN, Fig. 8A) which already demonstrated improved performance compared to state-of-the-art traditional methods [16]. Each layer of SRCNN consists of a convolution module followed by a nonlinear activation module. For natural images, 2D convolution kernels are routinely adopted, which are extended to 3D convolution kernels for MRI to harness the redundancy along an additional dimension for performance improvement [19, 23]. The kernel size dictates the network's receptive field, with larger kernels increasing the receptive field at the expense of more parameters. In practice, smaller kernels (e.g., 3×3×3) are typically chosen. A batch normalization module can be added after the convolution operation for easier training and faster convergence [73]. ReLU [74] and Parametric ReLU (PReLU) [75] are widely employed as activation modules due to their robust performance.

It has been demonstrated that augmenting the number of layers and introducing residual connections significantly enhances the performance of super-resolution networks [17] (Fig. 8B). The increased number of layers improves the network's capacity for nonlinear mapping, while residual connections alleviate training difficulties by simplifying the problem to learning the difference between high-resolution target and low-resolution input. The success of VDSR has spurred exploration of network architectures for further performance enhancement. DenseNet, which extensively connects all layers with numerous shortcuts [23, 34, 76] (Fig. 8C) was proposed to alleviate the vanishing-gradient problem, strengthen feature propagation, encourage feature reuse, and reduce the number of parameters. Another widely used network for image super-resolution is U-Net [77] (Fig. 8D), originally proposed for biomedical image segmentations. U-Net's multi-level architecture, utilizing max pooling layers in the downward path and upsampling in the upward path, along with shortcuts connecting within each level, allows it to capture both global information in the lower levels and local details in the upper levels. It demonstrates robust performance in various super-resolution applications [27, 67, 68, 78].

While task-specific modifications of network architectures can bring minor performance improvement [20, 25, 27, 34, 69, 79], widely-adopted representative networks such as VDSR and U-Net generally achieve robust and decent performance, serving as excellent starting points for most applications. When considering network architectures beyond basic CNNs, representative designs from previous studies also offer valuable references. For instance, for GANs, the discriminator in SRGAN [18] has proven useful in various applications [34, 36, 68]. Transformer architectures are more complex, integrating self-attention mechanisms, and often employ convolutional layers alongside transformer modules for efficient feature extraction. The original transformer [38] remains a solid reference for designing transformer modules. For diffusion models, U-Net is a popular choice for estimating the noise mean and variance at each time step [47-50]. Regarding INR, the encoder (CNN) and decoder (MLP) setup in [56] may provide insights for future designs.

4.2. Image evaluation metrics and loss functions

Another important aspect is the selection of the metrics for evaluating the similarity between super-resolved MRI and the target high-resolution image. Table 1 provides an overview of commonly employed evaluation metrics for image super-resolution. MSE, MAE, and peak SNR (PSNR) stand out as widely used metrics owing to their simplicity, clear physical interpretation, and simplicity of integration into mathematical optimization frameworks [80]. Despite their prevalence, these metrics, based purely on data consistency, are recognized as suboptimal indicators of image visual quality [18, 80, 81]. To address this limitation, structural similarity (SSIM) was introduced, aiming to better capture perceptual quality by considering luminance, contrast, and structural information within the images [80]. However, subsequent research has revealed a direct empirical and analytical link between SSIM and MSE-based metrics [82].

More recently, evaluation metrics derived from neural network-extracted features have been proposed, which attempt to better emulate human perceptions. The VGG perceptual loss, based on feature vectors from a VGG network pretrained on the ImageNet natural image dataset for object classification [83], is a notable example. Leveraging the extensive training samples in ImageNet, VGG perceptual loss is believed to capture visual features akin to those perceived by humans. Typically, feature vectors are extracted from deeper layers of the VGG network to encompass higher-level features [18]. Another metric, the Fréchet Inception Distance (FID), has been proposed for reference-free image quality assessment [84]. It quantifies the distance between feature vectors of real and network-generated images using the pretrained Inception v3 [85], another image recognition model. A lower FID score is indicative of higher visual quality.

| Evaluation metric | Calculation formular |
|---|---|
| Mean squared error (MSE) | $MSE = \|\|x_{HR} - \hat{x}_{SR}\|\|_2^2 = \frac{1}{N}\sum_{i=1}^{N}(x_{HR_i} - \hat{x}_{SR_i})^2$ |
| Mean absolute error (MAE) | $MAE = \|\|x_{HR} - \hat{x}_{SR}\|\|_1 = \frac{1}{N}\sum_{i=1}^{N}|x_{HR_i} - \hat{x}_{SR_i}|$ |
| Peak signal-to-noise ratio (PSNR) | $PSNR = 20\log_{10}\left(\frac{MAX_I}{\sqrt{MSE}}\right)$ |
| Structural similarity (SSIM) | $SSIM = \frac{(2\mu(x_{HR})\mu(\hat{x}_{SR}) + c_1)(2\sigma(x_{HR},\hat{x}_{SR}) + c_2)}{(\mu(x_{HR})^2 + \mu(\hat{x}_{SR})^2 + c_1)(\sigma(x_{HR})^2 + \sigma(\hat{x}_{SR})^2 + c_2)}$ |
| VGG perceptual loss (VGG loss) | $VGG\ loss = \|\|VGG(x_{HR}) - VGG(\hat{x}_{SR})\|\|^2$ |
| Fréchet inception distance (FID) | $FID = \|\|\mu - \mu_w\|\|^2 + tr(\Sigma + \Sigma_w - 2(\Sigma\Sigma_w)^{\frac{1}{2}})$ |

**Table 1. Image evaluation metrics.** Commonly used image evaluation metrics include mean squared error (MSE), mean absolute error (MAE), peak signal-to-noise ratio (PSNR), structural similarity (SSIM), VGG perceptual loss (VGG loss), and Fréchet inception distance (FID). $\|\|x_{HR} - \hat{x}_{SR}\|\|_1$ and $\|\|x_{HR} - \hat{x}_{SR}\|\|_2^2$ denote the L1 and L2 distance between $x_{HR}$ and $\hat{x}_{SR}$, respectively. $N$ is the total number of pixels inside the image. $MAX_I$ is the maximum possible value of the image. $\mu(x_{HR})$ and $\sigma(x_{HR})$ are the mean and variance of $x_{HR}$, respectively. $\sigma(x_{HR}, \hat{x}_{SR})$ denotes the covariance of $x_{HR}$ and $\hat{x}_{SR}$. $c_1 = (k_1 MAX_I)^2$ and $c_2 = (k_2 MAX_I)^2$ are the two variables to stabilize the division with weak denominator, where $k_1 = 0.01, k_2 = 0.03$ by default. $VGG(x_{HR})$ is the feature vector calculated from pre-trained VGG net with $x_{HR}$ as input. $\mathcal{N}(\mu, \Sigma)$ is the multivariate normal distribution estimated from Inception v3 features calculated on real life images and $\mathcal{N}(\mu_w, \Sigma_w)$ is the multivariate normal distribution estimated from Inception v3 features calculated on generated images.

Loss functions for network training are crafted based on image evaluation metrics, and by minimizing specific loss functions, the network can be optimized in terms of corresponding metrics. MSE-related loss functions are most popular choices thanks to the ease of their optimization and their ability to achieve optimized PSNR and data consistency, which however might introduce potential blurring to the generated images as discussed in Section 2.2. Recently, it is reported that minimizing MAE would lead to better image super-resolution performance compared to minimizing MSE [86]. SSIM-based loss functions have also been proposed, intending to produce images with improved SSIM [27, 86, 87]. Nevertheless, these images still exhibit visual similarities to those generated from MSE-related loss functions, given the direct relation between SSIM and MSE [82].

In the pursuit of better image visual quality, more advanced loss functions have been designed. GAN is a representative example which leverages adversarial loss to push synthesized images towards the manifold of target images, generating images with perceptually indistinguishable features such as contrast and textural details [18], as discussed in Section 2.1. In practice, a weighted summation of MSE-related loss and adversarial loss is often adopted to strike a balance between visual quality and data consistency (Eq. 5). Figure 9 visually illustrates the impact of adversarial loss, showcasing improved visual quality at the expense of reduced data consistency, as quantified by PSNR and SSIM (Fig. 9, iii) [35]. Similarly, VGG-based loss has proven effective in enhancing image visual quality while compromising data consistency [18].

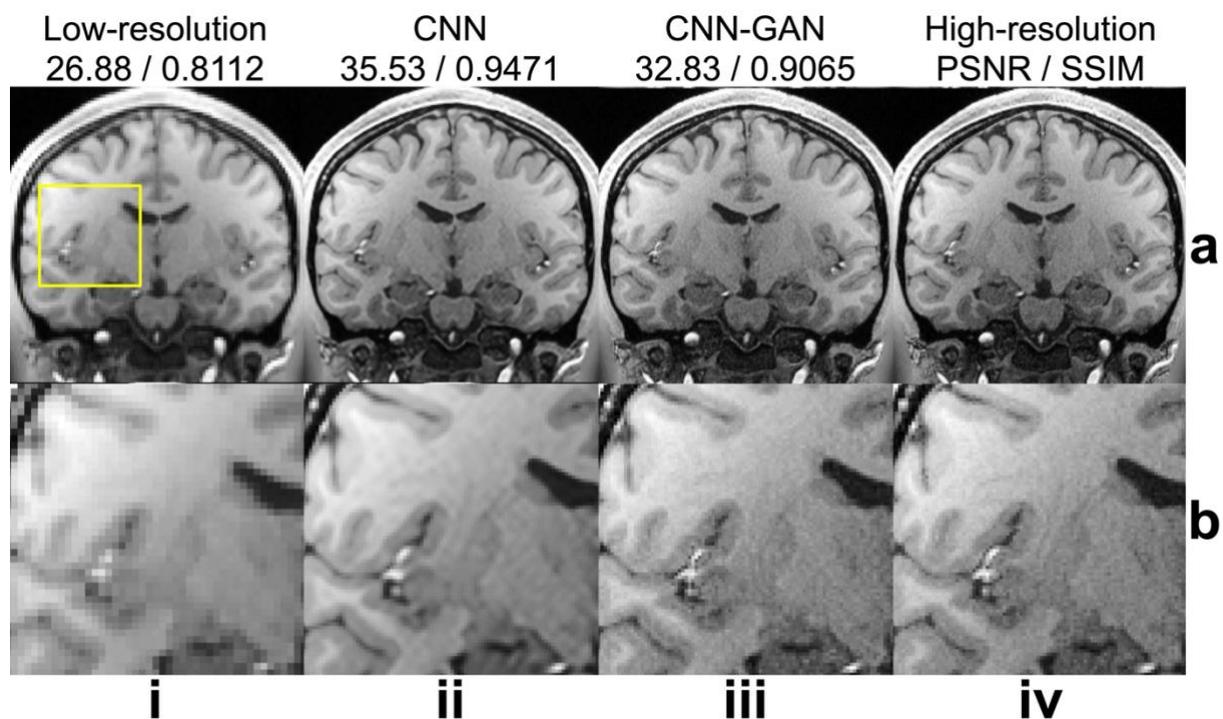

**Figure 9. Effect of adversarial loss.** The nearest neighbor-interpolated low-resolution (i), CNN trained with MAE loss super-resolved (ii), CNN trained with MAE and adversarial loss super-resolved (iii), and target high-resolution images (iv) (a) and an enlarged region (b) are demonstrated. The peak signal-to-noise ratio (PSNR) and structural similarity (SSIM) between the low-resolution / super-resolved images and the target high-resolution images are listed. The figure is reproduced from Fig.1 in [35] with permission.

In the context of MRI super-resolution, the selection of evaluation metrics and loss functions requires careful consideration. Beyond the output images themselves, downstream analyses and diagnoses are arguably more critical. As discussed in Section 3, comprehensive evaluation metrics tailored for specific clinical and neuroscientific applications have been proposed. Reader assessments by radiologists have been identified as supportive for the clinical value of deep learning-based super-resolved MRI [22, 27, 62, 63]. Furthermore, quantitative analyses of super-resolved images, such as brain region segmentation [67, 68] and cortical surface reconstruction [24, 70], provide valuable evidence of their benefits for neuroscientific research. Task-specific evaluations, derived from clinical and neuroscientific needs, are deemed more compelling than relying solely on traditional metrics like MSE and PSNR.

These task-specific evaluations can also offer insights into the selection of appropriate loss functions. Notably, it has been demonstrated that despite GAN-synthesized images exhibiting more visually appealing textures (as shown in Fig. 9), their quantitative analyses (e.g., brain region segmentation and cortical surface reconstruction) may not show improvement compared to those from CNN trained without adversarial loss. This discrepancy may be attributed to compromised data consistency introduced by the inclusion of adversarial loss [68]. The inclusion of adversarial loss in GANs also demands more training data and complicates network fine-tuning and generalization. Nevertheless, radiologists have expressed a preference for GAN-synthesized images with sharper and more realistic textures for brain MRI denoising [88]. Hence, the benefits of GANs for MRI super-resolution need to be carefully evaluated with respect to specific tasks.

Additionally, it is feasible to design tailored loss functions to generate images suitable for specific tasks. For instance, Iglesias et al. proposed the inclusion of a segmentation loss in the training of a super-resolution network for low-field brain MRI. This segmentation loss was calculated using a pre-trained 3D U-Net for brain region segmentation of high-resolution brain images. The incorporation of this segmentation loss enabled the network to generate high-resolution brain images with excellent brain segmentation accuracy amenable to automated quantitative morphometry [89].

4.3. Training data

The supervised deep learning methods discussed earlier necessitate paired low-resolution and high-resolution images for effective training. While such data can be prospectively acquired using low- and high-resolution protocols, this approach faces challenges such as long scan time and motion-induced inconsistency between the two scans. In light of this, we discuss an alternative approach that obtains training data through the retrospective downsampling of native high-resolution images, along with an examination of potential datasets suitable for method development and validation.

Downsampling strategies should ideally emulate the forward acquisition model of low-resolution images. For instance, to simulate thick-slice 2D acquisition (Fig. 1b), image-space downsampling is implemented as a 1D finite impulse response low-pass filter, mimicking a realistic RF profile [22, 63]. Alternatively, for simulating low in-plane resolution in 2D acquisition (Fig. 1a) or low isotropic resolution in 3D acquisition, k-space truncation can be utilized. This involves Fourier transforming the complex image data to k-space, cropping the central region, zero-padding, and then inversely Fourier transforming to obtain the interpolated low-resolution images [23, 27, 34, 35]. Additionally, low isotropic resolution images can also be obtained through image-space downsampling with anti-aliasing filtering and interpolation [24]. A recent advancement in achieving more realistic downsampling leveraged GAN to synthesize low-resolution images with textures indistinguishable from other low-resolution images from native high-resolution inputs [90].

The selection of public datasets for MRI super-resolution depends on the chosen downsampling strategy. For image-space downsampling, public datasets with high-resolution magnitude images are useful, including Human Connectome Project (HCP) [91], UK Biobank [92, 93], Alzheimer's Disease Neuroimaging Initiative (ADNI) [94], Massachusetts General Hospital Connectome Diffusion Microstructure Dataset (MGH CDMD) [95], etc. In contrast, k-space downsampling requires complex image data with both magnitude and phase information, which are typically found in datasets designed for MRI reconstruction, such as fastMRI [96].

5. Challenges and future directions

Deep learning-based MRI super-resolution has demonstrated promising performance potentially vauable for a variety of clinical and neuroscientific applications, but faces challenges in terms of the practicality, feasibility, and reliability. These challenges limit its broader adoption while offering potential directions for future research.

First of all, the practicality and feasibility of deep learning-based super-resolution are limited by the requirement of training data. The training of supervised learning methods necessitates paired low- and high-resolution images. Pham et al. used data from 21 subjects to train a 3D SRCNN with 3 layers and small number of parameters [19]. For a deeper network with more parameters such as VDSR, data from more subjects are usually involved in training (e.g., 64 training subjects in [24]). The inclusion of adversarial loss for GAN training brings larger training data requirement. Chen et al. incorporated over 1000 subjects for training a 3D GAN [34] presumably due to the difficulty for training a 3D discriminator which performs classification in a volume-by-volume manner. The training data requirement for GAN can be potentially reduced using a hybrid architecture with 3D generator and 2D discriminator, as demonstrated by Li et al. for 3D brain MRI denoising [88]. Data requirements for transformers and diffusion models are also significant (300 subjects for training a 2D transformer in [44] and 900 subjects for training a 2D diffusion model in [49]). While leveraging public datasets and retrospective downsampling, as discussed in Section 4.3, might address some of these issues, leveraging deep learning super-resolution models for tailored applications presents a significant challenge. Even for a relatively modest set of training data (e.g., 20 subjects), the associated time and financial costs for data acquisition are substantial.

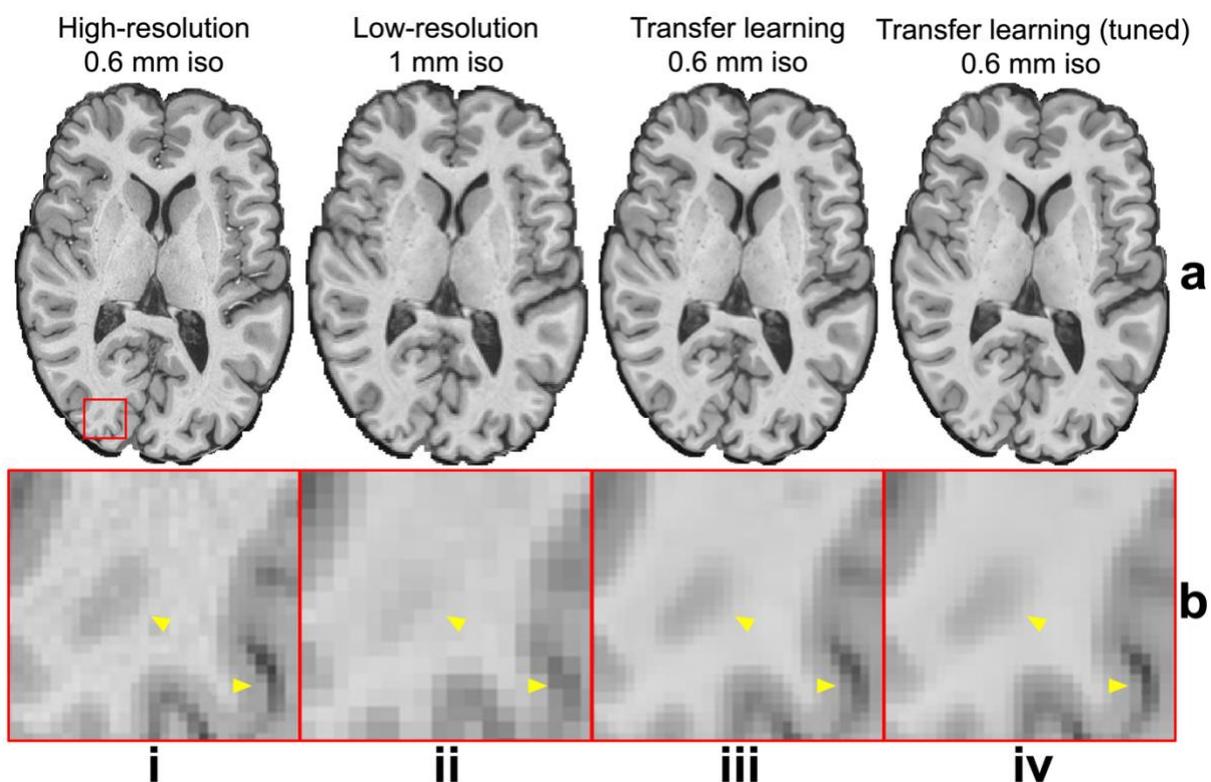

**Figure 10. Network generalization.** A representative axial slice (a) and an enlarged region (b) from native high-resolution (i), standard-resolution (b), and super-resolved (iii, iv) T1-

weighted images at 3T are demonstrated by Tian et al. [24]. The super-resolved images are obtained by directly applying a VDSR pretrained using 7T data (iii) and fine-tuning the pre-trained network using one subject of the 3T data (iv).

Transfer learning is a promising strategy to reduce the training data requirement. It is demonstrated that CNNs such as VDSR and U-Net is highly generalizable for MRI super-resolution [24, 68]. Tian et al. presented compelling evidence by applying a pre-trained VDSR, originally trained on 7T data, directly to 3T data, yielding satisfying results (Fig. 10, iii). Fine-tuning this pre-trained model with just one additional subject further enhanced the image sharpness (Fig. 10, iv) [24]. The decent generalizability suggests a promising approach – pre-training a super-resolution model on large-scale public datasets from HCP and UK Biobank and fine-tuning the pre-trained model using a relatively small dataset tailored to specific applications. This approach aligns with the concept of image quality transfer [97], postulating that rich information from high-quality images (e.g., high-resolution images from public datasets) can be effectively transferred to low-quality data (e.g., low-resolution images acquired for specific applications) using machine learning techniques. To date, the investigation of the transfer learning for more advanced deep learning-based super-resolution techniques beyond standard CNNs (e.g., GANs, transformers, diffusion models) is still limited, presenting a compelling direction for future exploration.

Self-supervised learning methods also offer a data-efficient solution, eliminating the need for external high-resolution training data. For instance, Zhao et al. introduced SMORE, a self-supervised approach for anisotropic resolution MRI with high in-plane but low through-plane resolution [98]. SMORE simulates low-resolution images from high-resolution in-plane slices and trains a model to super-resolve these simulated low-resolution images, which is applied to the low-resolution through-plane slices to improve their resolutions. This approach was extended to fetal MRI [99] and holds promise especially for modalities where 2D acquisitions are routinely adopted such as diffusion MRI and functional MRI.

INR introduced in Section 2.5 stands out as another low-data-demanding MRI super-resolution technique. By obtaining continuous image representations, INR enables mapping from continuous image coordinates to corresponding signals [55, 56]. Its coordinate-based design treats each pixel/coordinate as a training sample, coupled with straightforward encoder and decoder network architectures, making it trainable with limited data. Wu et al. successfully trained an INR for thick-slice brain MRI super-resolution with fewer than 10 training subjects, yielding decent image quality [59]. The potential of self-supervised INR for anisotropic MRI super-resolution has also been demonstrated, which constructs low- and high-resolution pairs from anisotropic MRI data, efficiently training the INR model [100]. The promising capabilities of INR to achieve high-quality super-resolution with a manageable amount of training data merit exploration in more neuroscientific and clinical applications.

Second, the reliability of deep learning-based methods is another hurdle for their wider adoptions. The inherent black-box nature of these algorithms raises concerns about the interpretability and trustworthiness of their outputs. Studies indicate that models trained with distribution matching losses (e.g., GANs) carry the risk of hallucinating features or even removing structures from the output images [101]. Given the super-resolution problem is intrinsically ill-posed, it is crucial to exercise caution to prevent such hallucinations, which could introduce ambiguity in scientific research conclusions and complicate clinical diagnostic, prognostic, and interventional decisions.

To enhance reliability, large-scale reader assessments of deep learning super-resolved MRI from clinicians are instrumental. To date, such assessments have been conducted for basic CNN-based super-resolutions [22, 27, 62, 63], supporting their use in clinical settings. The clinical adoption of more advanced super-resolution models, especially for those trained with distribution matching losses such as GANs and diffusion models, require more comprehensive evaluations from radiologists.

Efforts to introduce uncertainty quantifications have been made to bolster the robustness, safety, and interpretability of deep learning-based super-resolution. For example, Tanno et al. modelled the uncertainty in terms of intrinsic and parameter uncertainty [70, 71]. Intrinsic uncertainty reflects the inherent ambiguity linked to the ill-posed nature of the super-resolution problem, quantified through the variance of the target conditional distribution estimated from a separate network. Parameter uncertainty accounts for ambiguity in selecting model parameters, reduced using variational dropouts [102]. This approach captures various settings of network parameters for the given training data, providing a more robust result and an uncertainty estimation represented by the standard deviation. The efficacy of dropout layers for estimating uncertainty and improving robustness has been demonstrated in various applications, such as MRI reconstruction [103] and quantitative MRI [104]. A more thorough evaluation of the uncertainty estimation framework on advanced super-resolution models, along with a comprehensive assessment of generated uncertainty maps, especially in the context of patient data, may offer a promising avenue to further enhance the reliability and robustness of deep learning-based MRI super-resolution.

Finally, the ill-posed nature of super-resolution introduces the possibility of artifacts, hallucinations, and unrecoverable details during image translation [101]. This issue becomes particularly pronounced when discrepancies arise between the distributions of training and testing data, such as when applying models trained on healthy subjects to patient data. Currently, the demonstration and evaluation of such artifacts and their impact on downstream analysis remain limited. A thorough analysis of these artifacts could yield valuable insights into the potential risks associated with super-resolution and potential strategies for their mitigation. Furthermore, the partial volume effect inherent in low-resolution images presents an additional challenge in recovering fine details, even with advanced super-resolution methods. For instance, attempting to recover lesions smaller than 2 mm from a 2 mm resolution image would be nearly impossible. In such cases, an alternative approach may involve denoising high-resolution, low-SNR images [88, 105] rather than attempting to super-resolve low-resolution, high-SNR images. Both image denoising and super-resolution represent promising strategies for enabling accelerated MRI without significantly compromising image quality. However, further evaluations and discussions are warranted to determine their optimal use in various scenarios.

6. Conclusions

In conclusion, MRI super-resolution driven by deep learning techniques from CNNs to advanced models such as GANs, transformers, diffusion models, and INR models, holds great promise in revolutionizing clinical and neuroscientific applications. These techniques respond to challenges posed by limited MRI resolution, enhancing image quality, analysis, and diagnostic accuracy in research and clinical settings. Practically, classical architectures such as VDSR and U-Net prove robust. The use of non-data-consistency loss such as adversarial loss requires careful considerations especially for quantitative imaging tasks. Task-specific customization of loss functions and metrics is vital, extending assessment beyond traditional image metrics to downstream analyses. Despite these advancements, the feasibility and reliability of deep learning-based super-resolution MRI are confronted with several hurdles before it can be truly adopted broadly. Enhancing feasibility calls for innovative solutions like transfer learning and self-supervised learning to overcome limited training data. Ensuring reliability of AI-synthesized super-resolved MR images demands extensive assessments, robust uncertainty quantification, and comprehensive artifact analysis. While deep learning-based MRI super-resolution is rapidly evolving, this chapter provides a concise framework for navigating the current state of the field and identifies key avenues for future endeavours. Ongoing collaboration between researchers, clinicians, and neuroscientists is essential to unlock the full potential of AI-driven advancements in medical imaging.

Acknowledgement

The authors would like to thank Dr. Fredrick Lange, Minhao Hu, and Haoruo Cheng for their thoughts on specific techniques.